# Multi-Kernel Capsule Network for Schizophrenia Identification

Tian Wang, Anastasios Bezerianos, *Senior Member, IEEE*, Andrzej Cichocki, *Fellow, IEEE*, and Junhua Li*, *Senior Member, IEEE*

*Abstract— Objective:* Schizophrenia seriously affects the quality of life. To date, both simple (linear discriminant analysis) and complex (deep neural network) machine learning methods have been utilized to identify schizophrenia based on functional connectivity features. The existing simple methods need two separate steps (i.e., feature extraction and classification) to achieve the identification, which disables simultaneous tuning for the best feature extraction and classifier training. The complex methods integrate two steps and can be simultaneously tuned to achieve optimal performance, but these methods require a much larger amount of data for model training. *Methods:* To overcome the aforementioned drawbacks, we proposed a multi-kernel capsule network (MKCapsnet), which was developed by considering the brain anatomical structure. Kernels were set to match with partition sizes of brain anatomical structure in order to capture interregional connectivities at the varying scales. With the inspiration of widely-used dropout strategy in deep learning, we developed vector dropout in the capsule layer to prevent overfitting of the model. *Results:* The comparison results showed that the proposed method outperformed the state-of-the-art methods. Besides, we compared performances using different parameters and illustrated the routing process to reveal characteristics of the proposed method. *Conclusion:* MKCapsnet is promising for schizophrenia identification. *Significance:* Our study not only proposed a multi-kernel capsule network but also provided useful information in the parameter setting, which is informative for further studies using a capsule network for neurophysiological signal classification.

## I. INTRODUCTION

SCHIZOPHERNIA is among the most universal psychiatric disorders, affecting about 1% of the population worldwide [1], [2]. Patients with schizophrenia may experience difficulties in perception, emotions, and behaviors [3], [4]. At present, Schizophrenia diagnosis relies on the qualitative examination of obvious mental symptoms and patients' self-reported experience, which is not feasible to detect disease at the early stage. Machine learning-based diagnosis using neurophysiological signals might be able to detect subtle abnormality at the early stage of schizophrenia [5]-[10]. According to the findings of previous schizophrenia studies, functional dysconnectivity among disparate brain regions was repeatedly observed [11]-[14]. The functional dysconnectivity exhibited connectivity strength abnormalities between brain regions, which can be used to distinguish patients with schizophrenia from healthy people by machine learning methods [1], [15]-[22]. Up to now, both simple and complex methods have been employed for schizophrenia identification. For instance, Li et al. assessed all connectivity features to select top discriminative features and employed simple methods, such as linear discriminative analysis, to perform schizophrenia classification [22]. Other studies using complex methods (e.g., deep neural networks) achieved a better performance in schizophrenia classification according to the comparison results [1], [18]. However, these complex methods require a large amount of data for model training in order to reach such better performance. In practice, the scale of available data is usually not enough to meet the requirement due to a variety of factors including a limited number of participants and expensive cost in data collection.

Very recently, a new type of network called capsule neural network was proposed by Sabour et al., which does not require huge data for model training and could achieve good performance [23]. Capsule neural network was proposed to initially aim for classifying handwritten digits of postcodes and has now been extended to image recognition and text mining [24]-[30]. All these studies demonstrated that capsule neural network has advantages over other methods. For instance, capsule neural network outperformed convolutional neural network (CNN) in the recognition of brain tumour types based on the data of magnetic resonance imaging (MRI) [31]. Although the capsule neural network was very successful in the aforementioned applications, a good performance cannot be achieved when it is applied to schizophrenia identification without any adaptions and improvements. This is because schizophrenia is closely relevant to the brain module/community so that the anatomical structure in the brain have to be taken into consideration. To this end, we proposed a multiple-kernel capsule network, in which multiple kernels respectively correspond to varying parcellation sizes of the brain. For each kernel, the extracted connectivity information was represented as vectors, where the direction of a vector stands for an attribute while the length of the vector indicates the probability of being each attribute. These vectors were clustered by a routing-by-agreement algorithm to produce the final vectors representing predictions of each class.

* indicates the corresponding author (e-mail: juhalee.bcmi@gmail.com).

In this study, we first compared classification performances under different settings of the model to provide insights into how the performance was changed with them. Then, we compared the proposed model to the other methods which had been used for schizophrenia identification (i.e., k-nearest neighbours, k-NN; linear discriminant analysis, LDA; linear support vector machine, L-SVM; and deep neural network, DNN). A publicly available dataset was used to evaluate the performances of the methods and the performance was evaluated in terms of average classification accuracy obtained by the 10-fold cross-validation.

## II. METHODS

### A. Evaluation Dataset

All methods were evaluated using a publicly available dataset consisting of 148 participants, which is available at http://fcon_1000.projects.nitrc.org/indi/retro/cobre.html, provided by the Center for Biomedical Research Excellence [32]. High-resolution T1-weighted MRI and resting-state fMRI scans were collected by a 3-Tesla Siemens Trio scanner. The High resolution T1-weighted MRI was collected with the utilization of a multi-echo magnetization-prepared rapid gradient echo sequence (repetition time (TR) = 2.53 s, echo time (TE) = [1.64, 3.5, 5.36, 7.22, 9.08] ms, flip angle = 7°, slab thickness = 176 mm, field of view (FOV) = 256×256 mm, acquisition matrix = 256×256, voxel resolution = 1×1×1 mm³). The resting-state fMRI data were obtained by single-shot full k-space echo-planar imaging (EPI) with the inter-commissural line as a reference (TR = 2 s, TE = 29 ms, matrix size = 64×64, slices = 33, voxel resolution = 3×3×4 mm³).

### B. Data Preprocessing

The MRI data were preprocessed using three toolboxes: (1) statistical parametric mapping (SPM12), (2) resting-state fMRI data analysis toolkit [33], and (3) data processing assistant for resting-state fMRI advanced edition [34] in the environment of MATLAB (Mathworks, Inc., Natick, Massachusetts, USA). Three participants were excluded from the preprocessing procedure because of unavailable category information (there is no label to recognize whether the data were from a patient) or a short length of volume scanning, resulting in 145 participants. Additional 14 participants were removed due to excessive head movements. This exclusion resulted in 60 patients with schizophrenia and 71 healthy controls. After the preprocessing procedure comprising volume removal, motion correction, slice timing correction, spatial normalization, signal regression with the regressors of 24 head motion parameters, cerebrospinal fluid, and white matter, temporal band-pass filtering with cut-off frequencies of 0.01Hz and 0.08Hz and spatial smoothing, a parcellation atlas named automated anatomical labelling (AAL) was applied to parcellate the brain into 116 regions of interest (ROIs) [35]. Pearson correlation was subsequently utilized to estimate connectivity strengths for all pairs of ROIs. Fisher's r-to-z transformation was then applied to improve the normality of connectivity strength values. All values were assembled to form a functional connectivity matrix, representing as $M_{FC}$.

### C. Multi-Kernel Capsule Network

The functional connectivity matrix obtained after the pre-processing procedure was fed into the proposed deep learning model, namely the multi-kernel capsule network (MKCapsnet). Fig. 1 depicts the structure of the model. It consists of three layers: convolutional layer, capsule layer, and classification capsule layer. We set six kernels (i.e., kernel sizes: 1, 4, 6, 7, 9, and 15 columns) with diverse convolutional sizes in the first layer to match with varying region sizes of anatomical parcellation of the brain (e.g., a kernel size of 4 corresponds to the subcortical area and a kernel size of 9 corresponds to the cerebellum area). In the following layer (i.e., capsule layer), the extracted connectivity information from the first layer is represented as vectors (known as capsules), whose directions stand for attributes and whose lengths indicate the probabilities of being each attribute. These vectors are assigned to six channels corresponding to six kernels we set. With the inspiration of dropout strategy in deep learning, we set a capsule dropout strategy in the capsule layer, where the routing agreement algorithm is employed to learn based on capsules. Finally, the margin loss is utilized in the classification capsule layer to update weights by backpropagation process. The columns in the $M_{FC}$ were convoluted using kernels with different sizes. For each kernel, the output of the convolutional layer is a set of vectors (i.e., $u_i$ for capsule $i$) of the capsule layer. The vector $u_i$ is rotated and transformed into a predicted vector $\hat{u}_{j|i}$, which predicts the output of the capsule $i$ to higher level capsule $j$.

$$\hat{u}_{j|i} = W_{ij} u_i \tag{1}$$

where $W_{ij}$ is the weighting matrix updated by the backpropagation process. The input of capsule $j$ in the higher capsule level is the weighted summation of all the predicted vectors from the lower level capsules, obtaining by

$$s_j = \sum_{i \in I} c_{ij} \hat{u}_{j|i} \tag{2}$$

where $c_{ij}$ is a coupling coefficient, representing the routing coefficient from the lower level capsule $i$ to the higher level capsule $j$. $I$ is the set of all capsules in the lower level. The coupling coefficient $c_{ij}$ is determined by a softmax function as follows,

$$c_{ij} = \exp(b_{ij}) / \sum_{j \in J} \exp(b_{ij}) \quad (3)$$

where $b_{ij}$ is a logarithmic prior probability that capsule $i$ is coupled to capsule $j$, which is iteratively updated. $J$ is the set of all capsules in the higher layer.

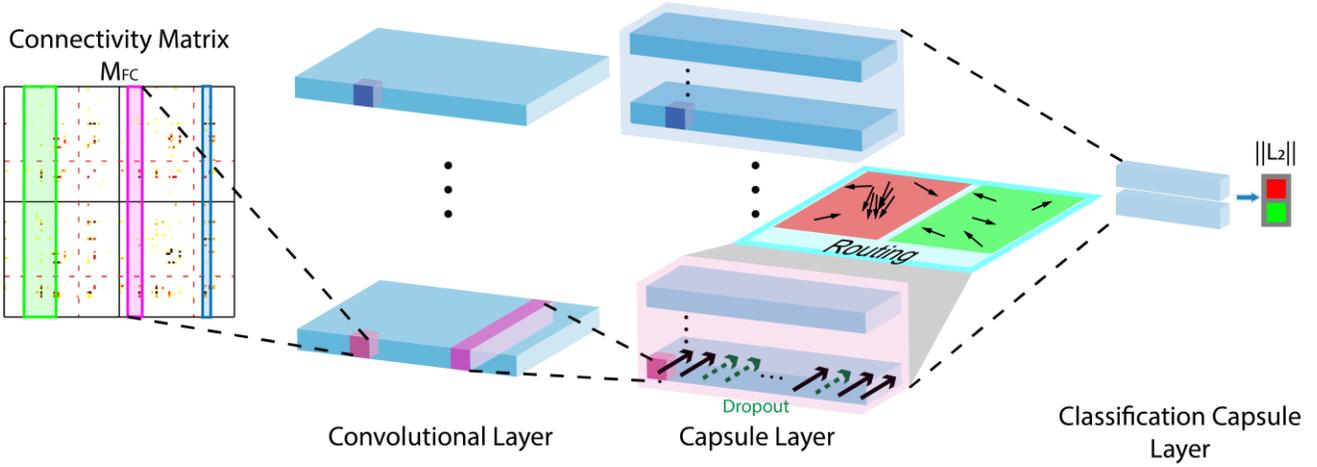

Fig. 1. Model structure of multi-kernel capsule network. The model consists of convolutional layer, capsule layer, and classification capsule layer. In the capsule layer, capsule dropout strategy was embedded inside each channel. The dropout was separately set for each channel and the dropout rate (50 %) was identical for all channels. Routing agreement algorithm was used to learn based on capsules.

$$b_{ij} \leftarrow b_{ij} + \hat{u}_{j|i} \cdot v_j \quad (4)$$

where $v_j$ is the output vector of capsule $j$ and obtained by a non-linear 'squashing' function as follows,

$$v_j = \frac{\|s_j\|^2}{1 + \|s_j\|^2} \frac{s_j}{\|s_j\|} \quad (5)$$

This step normalizes the length of the output vector to be within the range [0, 1]. The above procedure is called routing. After that, the outputs of the capsule layer are inputted into the subsequent classification capsule layer. The number of capsules in this layer is the same as the number of classes. During the testing phase, the lengths of the capsules are calculated to have probabilities of being each class. The class with the largest probability is the class of the sample. During the training phase, the total $L2$ margin loss is the sum of the losses of all capsules, calculating by

$$\begin{aligned} L2 &= \sum_{j \in J} L2_j \\ &= \sum_{j \in J} T_j \max(0, m^+ - \|v_j\|)^2 \\ &\quad + \sum_{j \in J} \lambda(1 - T_j) \max(0, \|v_j\| - m^-)^2 \end{aligned} \quad (6)$$

where $T_j$ is 1 if and only if the class $j$ is present. $m^+$ and $m^-$ are set as 0.9 and 0.1 respectively. The down-weighting factor $\lambda$ is set as 0.5.

## III. RESULTS

### A. Model Structure Comparison

TABLE I
PARAMETERS SETTINGS IN THE TRAINING

| Parameter | Setting |
|---|---|
| Epoch number | 500 |
| Learning rate | 0.01 |
| Batch size | 3 |
| Dropout rate | 0.5 |
| Early stop criterion | 0.008 |

Table I and Table II show the parameters used in the training and the parameters of layer settings used in the proposed model, respectively. As different settings in the model led to different classification performances, in this study, we compared performances in different settings (i.e., different dropout 4 strategies, single kernel (column) versus multiple kernels, with or without multi-slice channel, and different loss norms) to provide insights how the performance was changed with them. The comparison results were listed in Table III (all accuracies were obtained through 10-fold cross-validation). According to the results, the model with the multi-kernel, multi-slice channel, and capsule dropout strategy achieved the best performance (i.e., accuracy, 82.42%; sensitivity, 88.57%; specificity, 75.00%). With the benefit from the multi-kernel setting, the accuracy was elevated by 2.25% compared to that of the best column kernel setting (see rows 4 and 6 in Table III). It was dramatically improved by 17.69% compared to that of the square kernel setting (see rows 5 and 6 in Table III), which is frequently utilized in image and video processing when deep learning model is employed. Multi-slice channel was better than single slice channel in the performance. Moreover, the capsule dropout strategy performed 1.43% better than the scalar dropout (see rows 1 and 3 in Table III) and 0.60% better than the vector dropout (see rows 2 and 3 in Table III). In the performance comparison of loss norms, $L2$ loss showed obvious superiority

TABLE II
PARAMETERS OF LAYER SETTING USED IN THE PROPOSED MODEL

| Layer | Type | Kernel size | Stride | Filter/Slice size | Channel | Vector Length |
|---|---|---|---|---|---|---|
| 1 | Convolution | [{1, 4, 6, 7, 9, 15}*,108] | 1 | 64 filters | - | - |
| 2 | Capsule | [1, 1, 64] | 1 | 10 slices | 6 | 20 |
| 3 | Capsule | - | - | - | 2 | 20 |

* Kernel sizes {1, 4, 6, 7, 9, 15} in the convolution layer correspond to the anatomical brain parcellation.

TABLE III
PERFORMANCE COMPARISON AMONG DIFFERENT MODEL STRUCTURE

| Row | Dropout strategy | Kernel type | Multi-slice channel | Loss norm | Accuracy | Sensitivity | Specificity |
|---|---|---|---|---|---|---|---|
| 1 | Scalar` | Column(size 1) | ✗ | L2 | 77.14% | 78.93% | **75.00%** |
| 2 | Vector^ | Column(size 1) | ✗ | L2 | 77.97% | 84.82% | 70.00% |
| 3 | Capsule* | Column(size 1) | ✗ | L2 | 78.57% | 81.43% | **75.00%** |
| 4 | Capsule | Column(size 15) | ✗ | L2 | 78.63% | 83.04% | 73.33% |
| 5 | Capsule | Square(size 15) | ✗ | L2 | 63.19% | 71.61% | 53.33% |
| 6 | Capsule | Multiple | ✗ | L2 | 80.88% | **88.57%** | 71.67% |
| 7 | Capsule | Multiple | ✓ | L1 | 69.34% | 75.89% | 61.67% |
| 8 | Capsule | Multiple | ✓ | L2 | **82.42%** | **88.57%** | **75.00%** |

` Scalar Dropout: the dropout was performed on the elements of vectors.
^ Vector Dropout: the dropout was not separately set for each channel and a dropout rate of 50% was applied to the channels together
* Capsule Dropout: the dropout was separately set for each channel and the dropout rate of 50% was identical for all channels.

compared to $L$1 loss.

### B. Method Comparison

The proposed multi-kernel capsule network (MKCapsnet) was compared to other methods which had been used for the classification of schizophrenia based on functional connectivity [18], [22]. MKCapsnet outperformed all the other methods in terms of classification accuracy and sensitivity (see Fig. 2). MKCapsnet achieved the highest accuracy of 82% whereas k-Nearest Neighbours (k-NN), Linear Support Vector Machine (L-SVM), Linear Discriminant Analysis (LDA), and Deep Neural Network (DNN) performed accuracies of 71%, 73%, 76%, and 79%, respectively. In terms of sensitivity, MKCapsnet was at least 6% better than the others (MKCapsnet: 89% vs. k-NN: 57%, L-SVM: 67%, LDA: 70%, DNN: 83%). In terms of specificity, MKCapsnet was slightly worse than the other methods (MKCapsnet: 75% vs. k-NN: 83%, L-SVM: 77%, LDA: 82%, DNN: 75%).

Given that the routing is critical for the capsule network and it is valuable to look into details, we visualized the dynamic process of the routing in Fig. 3. The subplots in the first and second rows depict the evolution of $c_{ij}$ for the samples of patients with schizophrenia while the subplots in the third and fourth rows are for the samples of healthy controls. A value of

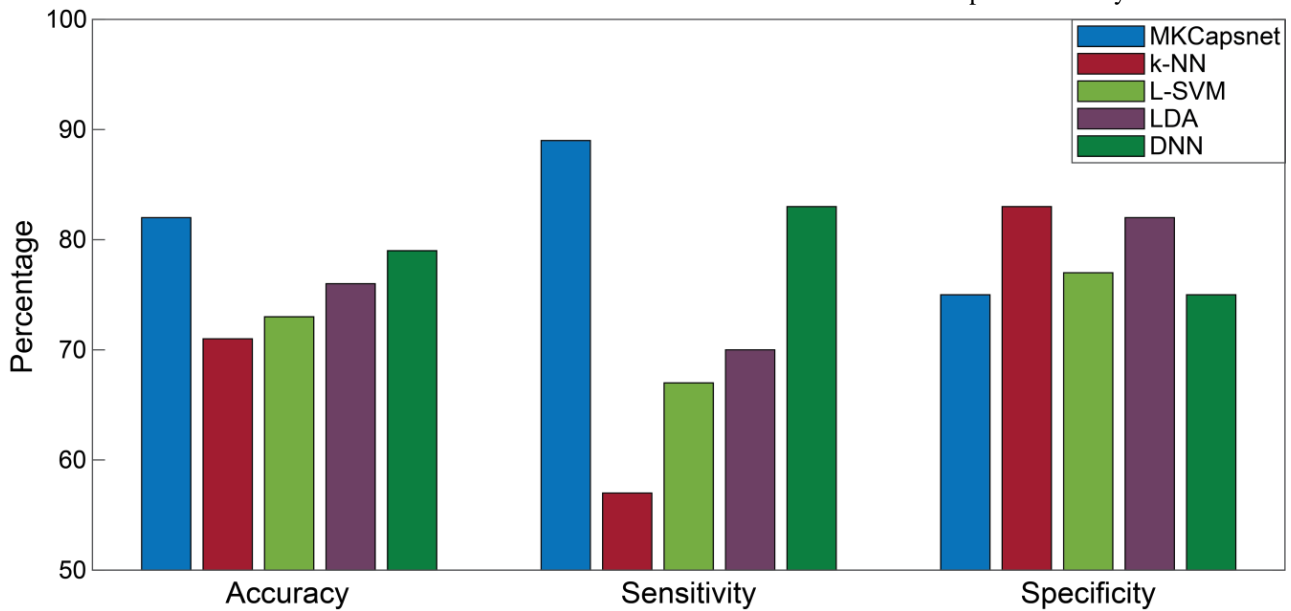

Fig. 2. Performance comparisons between the proposed model and k-Nearest Neighbours (k-NN), Linear Support Vector Machine (L-SVM), Linear Discriminant Analysis (LDA), and Deep Neural Network (DNN) methods in schizophrenia identification. For k-NN, L-SVM, LDA, and DNN methods, a feature selection procedure was utilized before the classification to boost performance as used in Li et al.'s paper [22]. The parameters used in the DNN model complied with the Kim et al.'s paper (3 hidden layers, 50 hidden nodes for each layer, and without pre-training) [18].

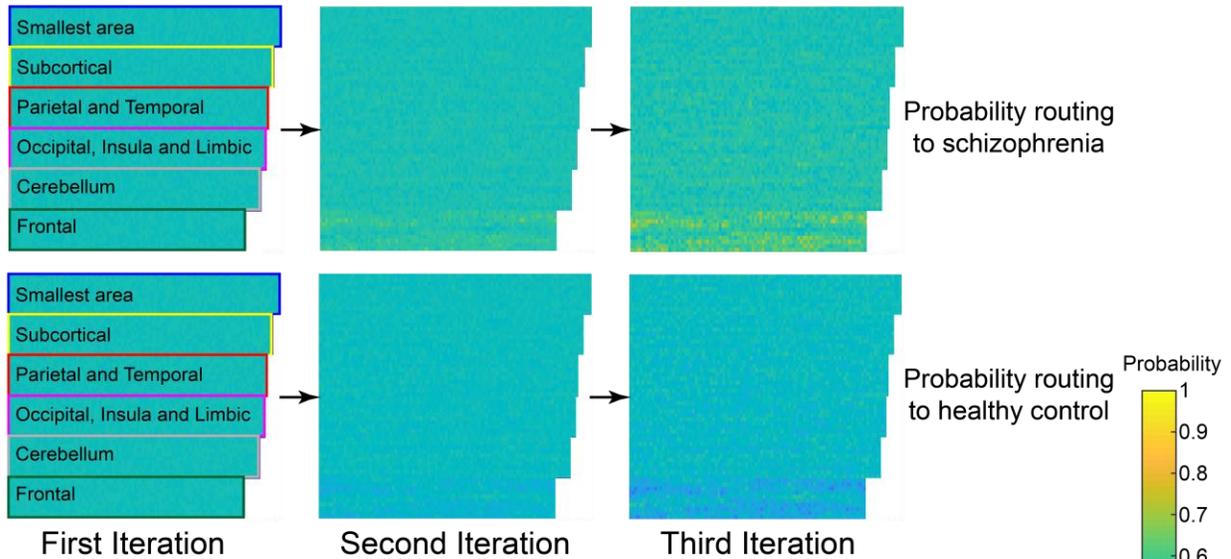

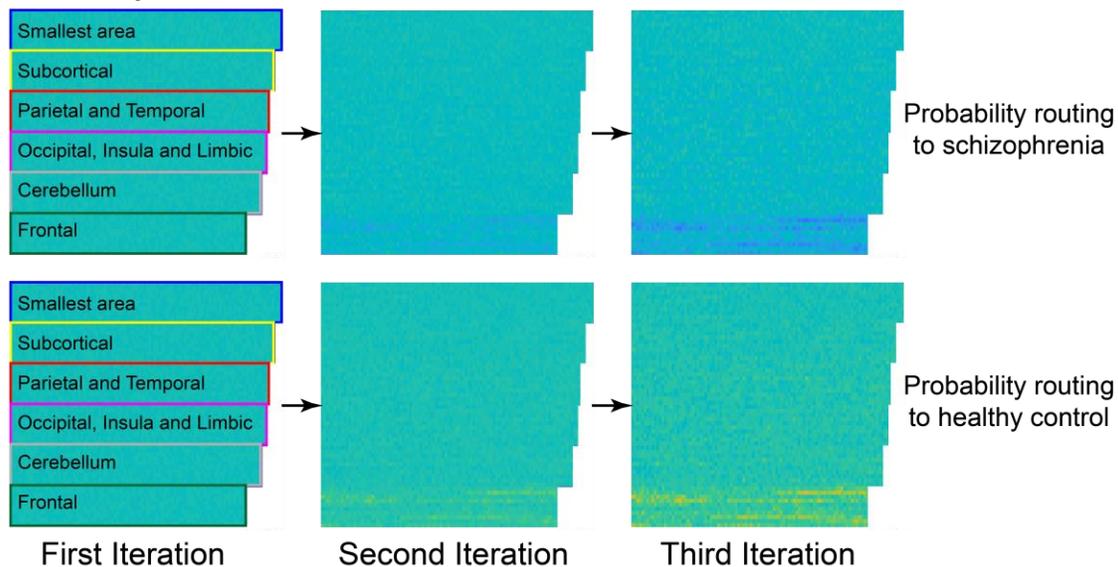

Fig. 3. The visualization of the routing process. The subplots in the first and second rows depict the evolution of $c_{ij}$ for the samples of patients with schizophrenia while the subplots in the third and fourth rows are for the samples of healthy control.

0 was assigned to initialize all $b_{ij}$. At the first iteration, $c_{ij}$ was calculated by "(3)" and was equal to 0.5. This value of 0.5 means that there is no preference to any class. With the evolution of $c_{ij}$, the paired values were gradually routed to 1 and 0, representing probabilities of being each class. As shown in Fig. 3, discriminative features (a pixel stands for one feature) exhibited high probability routing to the class of schizophrenia and low probability routing to the class of healthy control for the samples of patients with schizophrenia. In contrast, the probabilities routing to the classes were opposite for the samples of healthy control. Those features which were routed to the larger probability difference were more discriminative (showing dark yellow in the first and fourth rows and dark blue in the second and third rows in Fig. 3). We can see that all areas contributed to the schizophrenia identification but the frontal area contributed more than the other brain areas.

## IV. DISCUSSION

This study proposed a multi-kernel capsule network to identify schizophrenia disease using functional connectivity features. In this model, multiple kernels were embedded to capture intrinsic connectivity characteristics of varying sizes of anatomical brain areas. The comparison results demonstrated that the MKCapsnet overall outperformed the other methods which had been used for schizophrenia identification (i.e., k-NN, L-SVM, LDA, and DNN). In particular, the performance of MKCapsnet was 6% higher than that of the second best method in terms of sensitivity. This means that the MKCapsnet is able to more accurately identify patients with schizophrenia. In practical implication, it is the lower probability in the failure of schizophrenia detection in the case of a person with schizophrenia. In addition, the proposed method does not require an individual step of feature selection as used in methods such as k-NN and L-SVM. This reduces the number of steps for the classification procedure. The drawback of separate steps of classification procedure is that the feature extraction and classifier learning cannot be simultaneously tuned, which lowers the likelihood of the best optimization so as not to reach maximum performance.

We proposed multiple kernels to capture functional connectivity features of varying sizes of anatomical brain areas. This enables the model to have the capability to learn discriminative information existing in different scales from the local community to the global community. As demonstrated in the study, neither the smallest kernel size of 1 nor the largest kernel size of 15 performed the best performance. This might be because none of them can capture entire information existing in both small and large scales. This issue was tackled by the proposed multiple kernels. It is worth noting that the kernel sizes we set matched with the anatomical brain parcellation, rather than random selection. That is, a kernel size of 1 corresponds to the smallest area and a kernel size of 15 corresponds to the frontal area (kernel sizes of 4, 6, 7, and 9 correspond to the subcortical; pariental and temporal; insula, limbic and occipital; cerebellum respectively) when the whole brain was parcellated according to the AAL atlas [35]. A square kernel was frequently used in the image or video processing when deep learning model was utilized for classification or segmentation. This is not suitable when applying to our case because the region sequence was rearranged when assembling all connections into a matrix, which destroyed the original spatial relationship between regions. Therefore, we used the kernels including the entire column of the connectivity matrix so that all connections from one region to all other regions can be included. The rearrangement only affects the order of regions in each column. A kernel including the entire column is invariable to the inclusion of the regions. The results in our study showed that this kernel was better than the square kernel. In the future, 3D convolution could be employed so that connections can be assembled into a third-order tensor which retains the original spatial relationship between brain regions.

Based on the comparison results of different norms used in the loss function, the performance is differential. In our case, L2 loss was better than L1 loss. All three assessment indicators showed superior accuracy when the L2 loss was utilized (L2 vs.L1, accuracy, 82.42% vs. 69.34%; sensitivity, 88.57% vs. 75.89%; specificity, 75.00% vs. 61.67%). L1 loss exhibited fluctuant during searching an optimal solution and was less convergent. In contrast, L2 loss showed relatively stable convergence. In the capsule layer, we brought capsule dropout strategy to improve training effectiveness. Other than the scalar dropout strategy used in the image or video processing, we randomly discarded vectors. Moreover, we separately set the dropout rate for each channel (corresponding to each kernel) so that the number of vectors discarded in each channel can be kept identical. Compared to the vector dropout strategy (the dropout was not separately set for each channel and the number of vectors discarded in one channel might be more than that of another channel.), the classification performance was improved by 0.6 % when the capsule dropout strategy was used. The improvement was 1.4 % when compared to the scalar dropout scalar strategy which has been widely utilized in the deep learning models when processing the image or video data. These results demonstrated that the manner discarding entire vectors was better than that of discarding elements of a vector in the capsule network. The separate dropout in each channel gives the advantage of that the dropout rate would not imbalanced across channels. Therefore, it avoids that there is excessive dropout in some channels whereas there is a lack in the others.

## V. Conclusion

In this study, we proposed a multi-kernel capsule network to identify schizophrenia and provided detailed performance comparisons in terms of model structure and method. These results could give heuristic cues for further studies as our study is the first attempt to identify schizophrenia based on functional connectivity by a capsule network. Due to that it is at a very early stage to develop a capsule network for disease detection, there is a large space to improve performance from many angles. For instance, vector representation in the capsule network can be replaced by tensor representation. In this case, additional information can be represented besides the direction and probability that have been represented by a vector. In summary, our study demonstrated that capsule network was feasible and promising in the identification of schizophrenia. This model can also be extended to detect other diseases after appropriate adaption. Further efforts are required to improve the performance and broaden applications of the capsule network.